\documentclass[letterpaper, 10 pt, conference]{ieeeconf}

\IEEEoverridecommandlockouts                              
                                                          
\usepackage{graphicx}
\usepackage{multirow}
\usepackage{url}
\usepackage{amsmath}
\usepackage{amsfonts}
\usepackage{color}
\usepackage{colortbl}
\usepackage{xcolor}
\usepackage{soul}
\usepackage{tikz}
\usetikzlibrary{arrows}
\usepackage[labelformat=simple]{subcaption}

\begin{document}

\setlength{\abovedisplayskip}{3pt}
\setlength{\belowdisplayskip}{3pt}

\title{Focal Loss Dense Detector for Vehicle Surveillance}

\author{Xiaoliang~Wang, Peng~Cheng, Xinchuan~Liu, Benedict~Uzochukwu%
\thanks{X. Wang, P. Cheng, X. Liu and B. Uzochukwu are with the Technology Department, College of Engineering and Technology, Virginia State University, Petersburg, VA, 23806 USA e-mail: \{xwang,pcheng,xliu,buzochukwu\}@vsu.edu.}%
}


\maketitle

\begin{abstract}
Deep learning has been widely recognized as a promising approach in different computer vision applications. Specifically, one-stage object detector and two-stage object detector are regarded as the most important two groups of Convolutional Neural Network based object detection methods. One-stage object detector could usually outperform two-stage object detector in speed; However, it normally trails in detection accuracy, compared with two-stage object detectors. In this study, focal loss based RetinaNet, which works as one-stage object detector, is utilized to be able to well match the speed of regular one-stage detectors and also defeat two-stage detectors in accuracy, for vehicle detection. State-of-the-art performance result has been showed on the DETRAC vehicle dataset.
 
\end{abstract}

\IEEEpeerreviewmaketitle

\section{INTRODUCTION}
The traffic surveillance system is broadly used to monitor traffic conditions.
Vehicle detection plays significant roles in many
vision-based traffic surveillance applications. Vehicles need to be located based on the videos or images of the
traffic scene. Some further processing, such as vehicle tracking and vehicle counting, could be developed based on the obtained specified location of the vehicle. The extracted bounding box, which contains the vehicle, can also be collected for other usage, such as vehicle type and model recognition.
However, there are still a fairly amount of concerns with the development of vehicle detection technology.
One of them is the occlusion, which place resistance to accurate vehicle detection. In the real application scenario of traffic surveillance systems,
detection performance could also be influenced by different illumination and weather conditions. Vehicles and other objects could bring shadows, which easily give rise to false positives in detection procedures. Different vehicles may diversify in shape, size and color. Various pose may generate different appearance for the vehicle, which make vehicle detection even more challenging.
Previously, different feature extraction techniques have been employed for vehicle detection,relying on the rigid characteristic of vehicle\cite{dollar2014fast} and unique part based features\cite{felzenszwalb2010object}. Recently, Convolutional Neural Network(CNN)
has been proved to be a promising approach for feature extraction of Region of Interest in images. There are many CNN based methods, which have been proposed for vehicle detection and classification \cite{yang2015large,rujikietgumjorn2017vehicle,zhou2016image}. Nonetheless, superior detection accuracy and low processing time latency could hardly be achieved at the same time.  

In this study, we deploy a Focal Loss Convolutional Neural Network based object detection method-RetinaNet\cite{lin2017focal} to undertake the vehicle detection task for DETRAC\cite{wen2015detrac} dataset. Our experiment result show that the RetinaNet could be well adjusted to perform faster and more accurate vehicle detections compared to previous other methods.

\section{Overview of Focal Loss Dense Object Detector}

\subsection{Evolution of object detection method}

There are a group of traditional and classic object detection methods developed with history. Firstly, the sliding-window approach is proposed,
through which a classifier is applied on a dense image grid. Some of the representative work are the
study\cite{lecun1989backpropagation,vaillant1994original}, which utilize Convolutional Neural
Networks for handwritten digit recognition. The usage of boosted object detectors for face
detection has been explored in\cite{viola2001rapid}, which make the proposed methods widely accepted in the related area.
The study of integral channel features\cite{dollar2009integral} and HOG\cite{dalal2005histograms} lead to breakthrough for pedestrian detection.  
The method of DPMs\cite{felzenszwalb2010cascade} is able to make dense detectors applicable to general
object detection, which continuously achieve remarkable results on PASCAL\cite{everingham2010pascal}. However, with the revival of deep learning based methods for computer vision\cite{krizhevsky2012imagenet}, the traditional sliding-window approach was replaced by the unrivaled two-stage detectors, which dominate object detection lately.

For two-stage object detector, the Selective Search method\cite{uijlings2013selective} is the earlier work which utilize the first stage to generate sparse proposals which may potentially include objects inside and the second stage to classify the proposal as foreground or background. R-CNN\cite{girshick2014rich} is able to leverage Convolutional Neural Network for the second stage classification task and achieve even higher accuracy in object detection.  R-CNN make each object proposal in an image to pass through CNN independently for feature extraction, which lead to large time latency when executing object detection work. In order to accelerate, the SPPnet\cite{he2015spatial} and the Fast R-CNN\cite{girshick2015fast} pass through the CNN only once for the entire input image. For the further development of two-stage object detector, in the work of Faster R-CNN\cite{ren2017faster}, the first stage of Convolutional Neural Network(CNN) is used for generating Region of Interest(ROI) proposal; the second stage of CNN is used for both region proposal refining and object classification. The critical part is to make RPN share the full-image convolution features with the detection network. Based on the analysis of Faster R-CNN framework, many improvement work has been deployed\cite{shrivastava2016beyond,lin2016feature,dai2016r,he2017mask,dai2017deformable}.

For one-stage object detector, OverFeat\cite{sermanet2013overfeat} was one of the pioneered work based on deep networks.
Lately SSD\cite{fu2017dssd,liu2016ssd} and YOLO\cite{redmon2016you,redmon2016yolo9000} are the typical one-stage methods. In their study, Huang\cite{huang2016speed} discuss and analyze the accuracy and speed trade-offs among different CNN based object detectors. As their work analyzed,normally two-stage object detectors perform more accurate than one-stage object detectors; however, one-stage object detectors exhibit faster speed than two-stage object detectors. Until recently, one-stage detector RetinaNet\cite{lin2017focal} is able to achieve comparable accuracy as two-stage detector while still maintaining fast speed. 

Most one-stage detectors meet with the problem of class imbalance. The detectors usually go through a huge amount of location with only a few of them containing interested objects. Those easy negatives, which include little useful information, make training procedure rather inefficient; on the other side, the easy negatives would produce degenerate training models. Many study\cite{sung1996learning,viola2001rapid,felzenszwalb2010cascade,shrivastava2016training,liu2016ssd} employ hard negative mining methodologies to gain more information from hard samples within training procedures. Some more complicated sampling or reweighing methods are explored in \cite{rota2017loss}. Focal loss introduced in next part is proposed to solve the class imbalance issue.

\subsection{Focal Loss Dense Object Detector}

The normal Cross Entropy (CE) loss for binary classification is showed below:
\begin{equation}
  p_{t}=\begin{cases}
    p & \text{if $y=1$}\\
    1-p & \text{otherwise}
  \end{cases}
\end{equation}
  
\begin{equation}
CE(p,y) = CE(p_{t}) = -\log(p_{t})
\end{equation}

In the above equation, $y \in \{-1,1\}$ specifies the ground-truth class and
$p \in [0, 1]$ is the model’s estimated probability for the class
with label $y = 1$. As analyzed in \cite{lin2017focal}, this regular Cross Entropy loss function could easily be influenced by the sample imbalance of foreground and background class, which would unfortunately lead to instability in one-stage object detector training processes. Focal Loss function is proposed to solve this issue.

The Focal Loss could be defined as below.
\begin{equation}
FL(p_{t})=-\alpha_{t}(1-p_{t})^\gamma log(p_{t}) \\
\end{equation}
A weighting factor $\alpha \in [0,1]$  is incorporated for class 1 and $1-\alpha$ for class -1. As used in Cross Entropy(CE) loss,$p_{t}$ represent the estimated probability for class 1.The parameter $\gamma$ is used to control the speed at which easy examples are down weighted.
Previously, with default configuration, equal probability is given to binary classification to output either y = −1 or 1 when initialized.
In that case, because of the existence of class imbalance,the loss generated by proportionally dominant class would contribute more to the loss and lead to failing to converge in the training. So in order to further prevent the instability in training, a ‘prior’ variable $\pi$ is introduced, through which the value of $p$ estimated by the model for the rare foreground class could be set low,such as 0.01,at the beginning of training.This pre-configuration method could help system avoid diverging in training. 

\begin{table}[tbp!]
	\caption{Varying $\gamma$ and $\alpha$ for Focal Loss}
	\vspace{-5pt}
	\centering
    	\begin{tabular}{c|c|c}
    	\hline	
	$\gamma$ & $\alpha$ & mAP \\ \hline	
        0 &0.75& 69.63\\ \hline
        0.1 &0.75&69.92 \\ \hline
        0.2 &0.75&70.28 \\ \hline
        0.5 &0.5& 71.24\\ \hline
        1.0 &0.25& 71.85\\ \hline
        2.0 &0.25& \textbf{72.38} \\ \hline
        5.0 &0.25&70.87 \\ \hline
          	\end{tabular}
	\label{tab:vehicleDetectionCompare1}
	\vspace{-5pt}
\end{table}

\begin{figure}[tbp!]
	\centering
    
    \begin{subfigure}[b]{0.48\textwidth}
	\includegraphics[width=\textwidth]{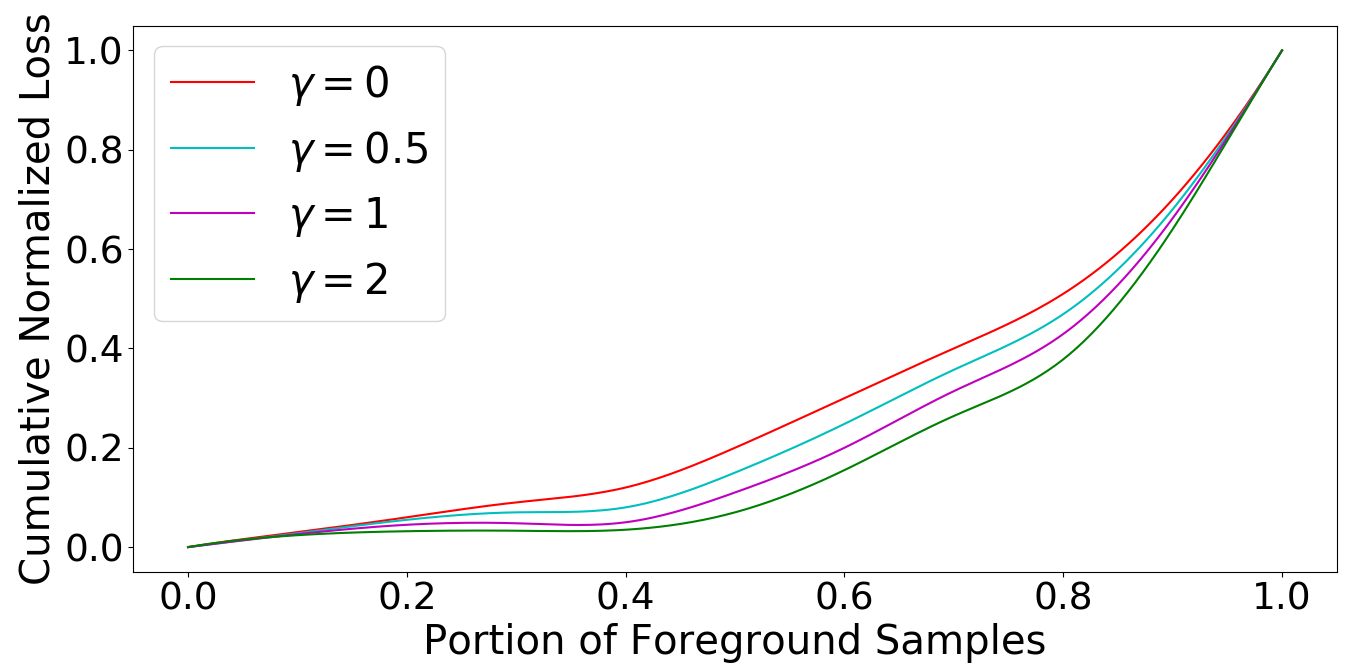}
    \caption{Foreground}\label{fig:foreground}
    \end{subfigure}    
    
    \begin{subfigure}[b]{0.48\textwidth}
    \includegraphics[width=\textwidth]
    {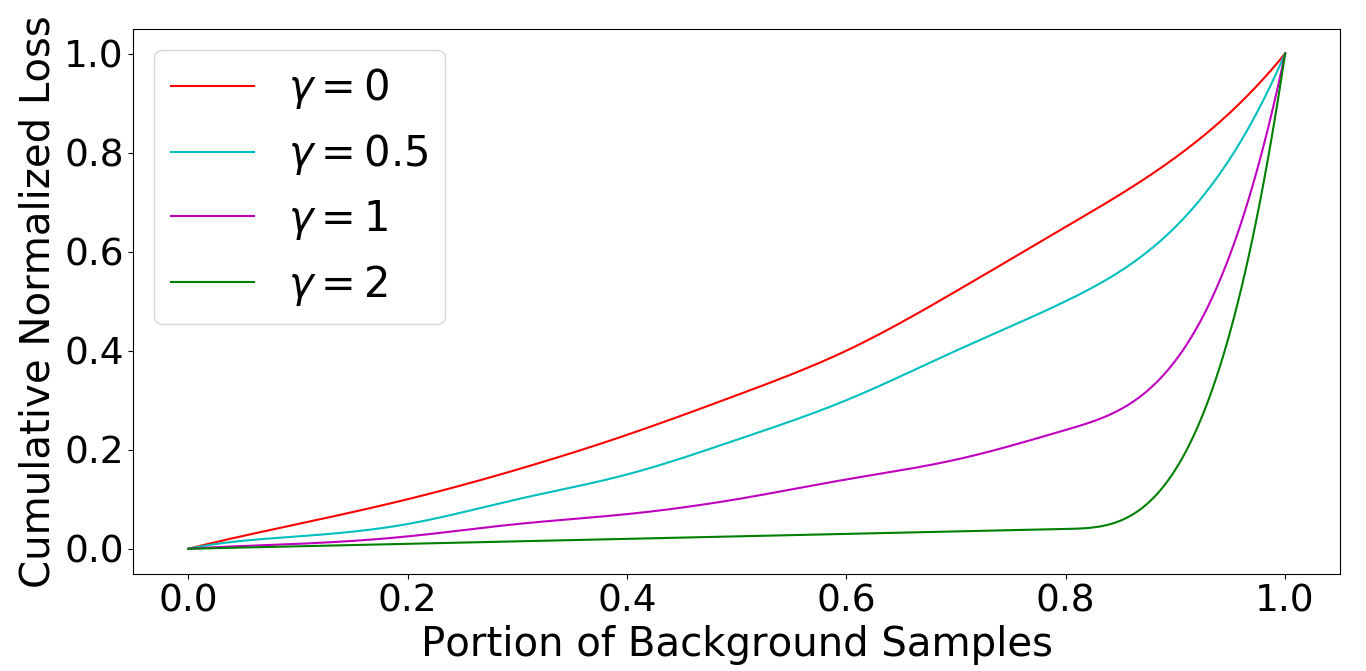}
    \caption{Background}\label{fig:background}
    \end{subfigure}
    
	\caption{Cumulative distribution functions of the normalized loss for positive and negative samples for different values of $\gamma$ for a converged
model.}
	\label{fig:lossresult}
	\vspace{-10pt}
\end{figure}

\begin{table*}[tbp!]
	\caption{Accuracy and Speed Result on DETRAC dataset.($ \gamma = 2.0, \alpha = 0.25$)}
	\vspace{-5pt}
	\centering
    	\begin{tabular}{c|c|c|c|c|c|c|c}
    	\hline	
	method & category  & FPS & mAP & car & van  & bus & others \\ \hline	
         SSD(ResNet-50)&one-stage& 22.74& 65.36&79.64&66.34&86.43&29.04 \\ \hline
         Faster R-CNN&\multirow{2}{*}{two-stage}& \multirow{2}{*}{12.21}&\multirow{2}{*}{71.93}& \multirow{2}{*}{88.57} & \multirow{2}{*}{76.42} & \multirow{2}{*}{90.09} & \multirow{2}{*}{32.63} \\ 
         (ResNet-50)  &              & &&&& \\ \hline
         RetinaNet(ResNet-50)&one-stage& \textbf{23.37} & 72.38& 89.01& 76.87& 90.49 & 33.15 \\ \hline
         SSD(ResNet-101)&one-stage& 20.54& 66.88 & 81.06& 67.76& 88.15 &30.56 \\ \hline
         Faster R-CNN&\multirow{2}{*}{two-stage}&\multirow{2}{*}{10.18}&\multirow{2}{*}{73.27}& \multirow{2}{*}{89.92} &\multirow{2}{*}{77.82}&\multirow{2}{*}{91.44} &\multirow{2}{*}{33.91} \\ 
         (ResNet-101)  &              &&&&& \\ \hline
         RetinaNet(ResNet-101)&one-stage&21.23& \textbf{73.79}& \textbf{90.43}& \textbf{78.35}& \textbf{91.86}&\textbf{34.52} \\ \hline
          	\end{tabular}
	\label{tab:vehicleDetectionCompare}
	\vspace{-5pt}
\end{table*}

\begin{figure*}[t!]
	\centering   
    
    \begin{subfigure}[b]{0.48\textwidth}
	\includegraphics[width=\textwidth]{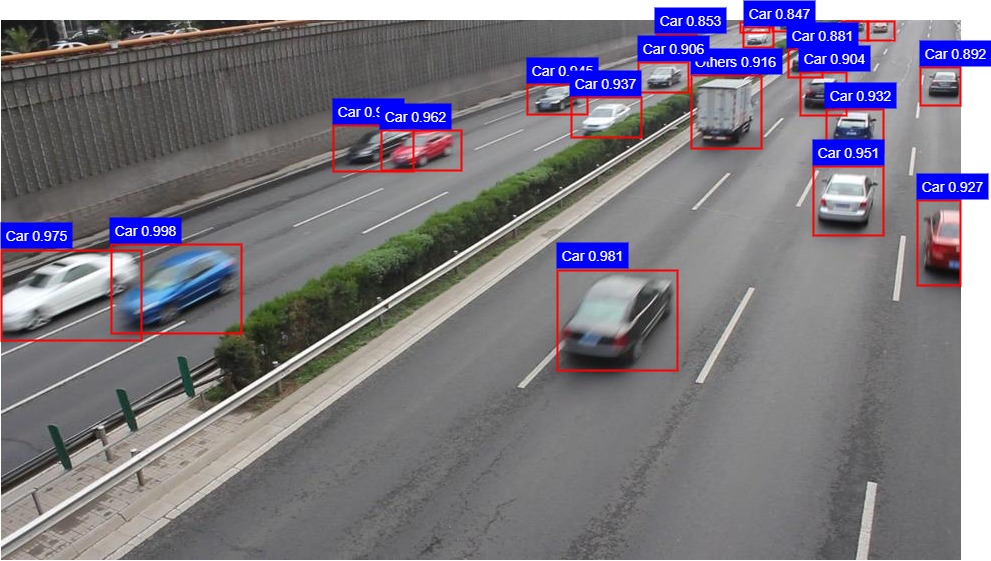}
    \caption{Daytime}\label{fig:daytime}
    \end{subfigure}    
    \begin{subfigure}[b]{0.48\textwidth}
    \includegraphics[width=\textwidth]
    {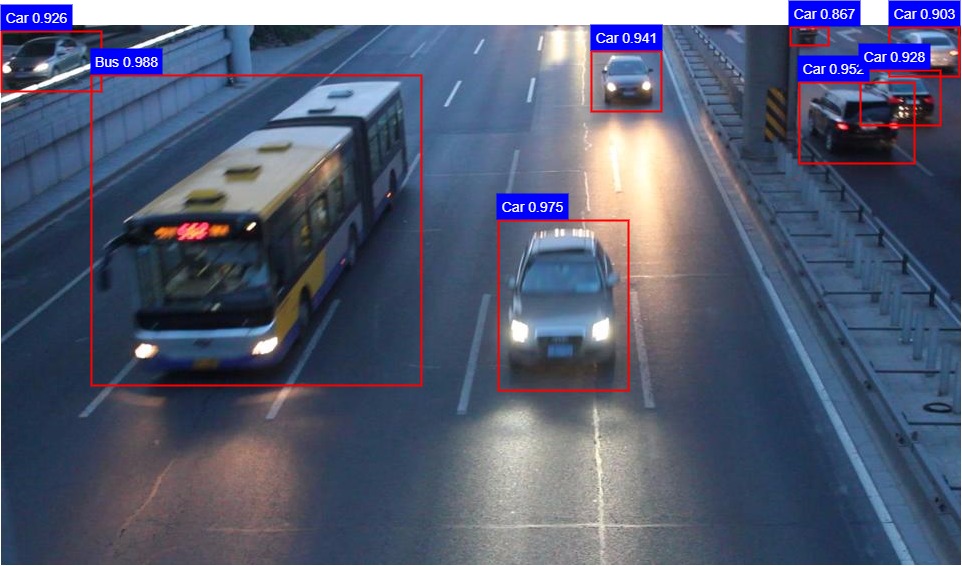}
    \caption{Nighttime}\label{fig:night}
    \end{subfigure}    
    \begin{subfigure}[b]{0.48\textwidth}
    	\includegraphics[width=\textwidth]{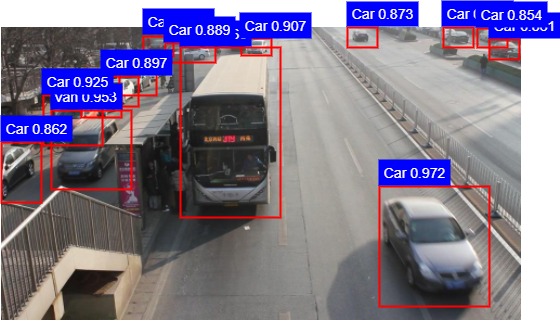}
        \caption{Sunny Weather}\label{fig:sunny}
        \end{subfigure}        
        \begin{subfigure}[b]{0.48\textwidth}
    \includegraphics[width=\textwidth]
    {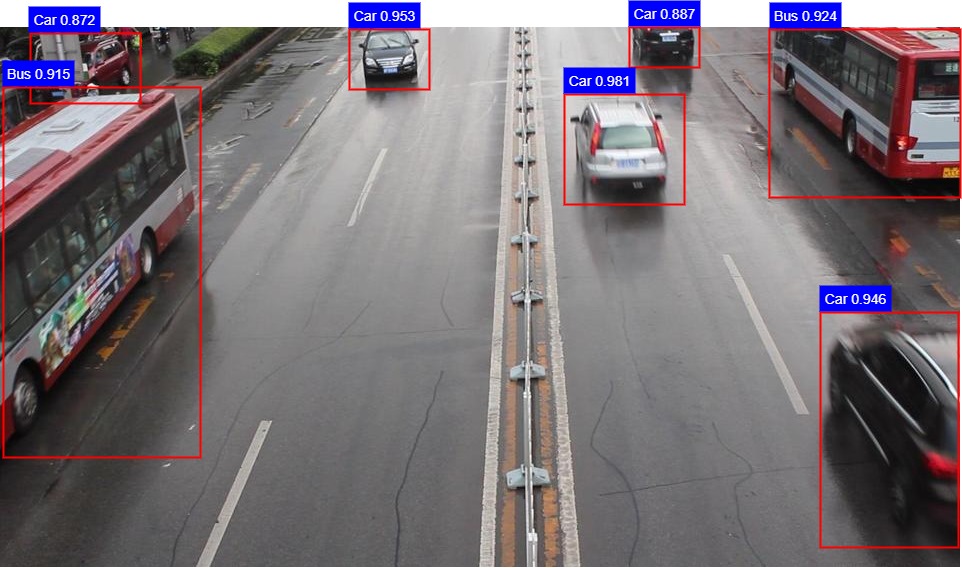}
    \caption{Rainy Weather}\label{fig:rainy}
    \end{subfigure}
   
	\caption{RetinaNet based Vehicle Detection Result on DETRAC Dataset}
	\label{fig:detectionResult}
	\vspace{-10pt}
\end{figure*}

\section{EXPERIMENT}

The dataset we use for experiment is DETRAC\cite{wen2015detrac} vehicle dataset. The dataset includes video taken in both daytime and night. It contains different weather conditions, such as sunny, cloudy and rainy situations. Four vehicle categories are defined in the dataset, which are \textit{car},\textit{bus},\textit{van} and \textit{others}(trucks and vehicles with trailers are categorized into \textit{others} group). The algorithm we use is the RetinaNet proposed in \cite{lin2017focal}. RetinaNet is able to match the speed of previous
one-stage detectors while surpassing the accuracy of
all existing state-of-the-art two-stage detectors on MSCOCO dataset\cite{lin2014microsoft}. The RetinaNet network architecture uses a Feature Pyramid Network (FPN) \cite{lin2016feature} backbone on top of a feedforward
ResNet architecture \cite{he2016deep} to generate a rich, multi-scale convolutional feature pyramid. The base ResNet models are pre-trained on ImageNet. For the final convolutional
layer of the classification subnet, we set the bias initialization
to $b= -log((1-\pi)/\pi)$, $\pi = 0.01$ in our experiments. As
explained previously, this initialization strategy prevents the large amount
of background anchors from generating a large, diverging
loss value in the training.

The code was implemented with MXNet\cite{chen2015mxnet} and run on a server equipped with two 
Intel 10-core Xeon CPU E5-2630 and an NVIDIA Tesla
K80 GPU. In our experiment, RetinaNet is trained with stochastic gradient
descent (SGD). Unless otherwise specified, all models are trained for 110k iterations with an initial learning rate of 0.01, which is then
divided by 10 at 70k and again at 90k iterations. We use
horizontal image flipping as the only form of data augmentation
unless otherwise noted. Weight decay of 0.0001 and
momentum of 0.9 are used.

Focal loss has been used as the loss on the output of the classification subnet in RetinaNet. Heuristically, we find the parameter setting $\gamma \in [0.5,5]$ and $\alpha \in [0.25,0.75]$ robust for RetinaNet. We choose $\gamma = 2$, $\alpha = 0.25$ to work best for our experiment, which is showed in Table \ref{tab:vehicleDetectionCompare1}.

To explore the effect of the focal loss
further, an experimental analysis is provided towards the distribution of the loss for a
converged training model. For the training configuration, we select RetinaNet with ResNet101
architecture and set the parameter $\gamma = 2$ (which obtained 73.79
mAP). We deploy this model randomly to a great amount of testing images
and record the predicted probability for around $10^6$ negative samples and $10^4$ positive samples. We collect the focal loss for those negative and positive samples
and normalize the sum of loss for each group to one. The
normalized loss is then sorted from low to high to obtain Cumulative Distribution Functions(CDF).
CDF for positive and negative
samples with different settings of $\gamma$ are shown in Figure~\ref{fig:lossresult}. According to the foreground samples result showed in Figure~\ref{fig:foreground}, it could be found that various settings of $\gamma$ has minor effect on CDF. Around 18\% of
the hardest positive samples occupy roughly half of the
positive loss.With $\gamma$ increasing, more of the loss gets focused
in the top 18\% of examples, but the influence is trivial.
However, as showed in Figure~\ref{fig:background},the various settings of $\gamma$ affect negative samples significantly.
For $\gamma=0$, the positive and negative CDFs looks fairly
similar. But with the value of $\gamma$ becoming larger, considerably more weight has been placed on the hard negative samples. It is showed that, with $\gamma=2$ (our training setting), the broad majority of
the loss is generated from a small portion of examples. This could help prove that focal loss can attenuate the impact from easy negatives,
transferring all attention to the hard negative samples.

In order to compare with previous work, for network architecture, we deployed 3 methods: SSD, Faster R-CNN and RetinaNet. As introduced previously,SSD and RetinaNet work as one-stage object detectors.Faster R-CNN works as a two-stage object detector.We chose either 50 or 101 for the ResNet depth. The accuracy and speed result are showed in Table \ref{tab:vehicleDetectionCompare}. We can find that the focal loss based RetinaNet could achieve higher accuracy than the representative two-stage object detector-Faster R-CNN. In addition, RetinaNet is able to run much faster than the two-stage object detector in terms of inference speed. Figure~\ref{fig:detectionResult} depicts RetinaNet detection result on DETRAC dataset under different illumination conditions and weather situations.Figure~\ref{fig:daytime} and Figure~\ref{fig:night} show the detection result in different periods of the day,which reflect different lighting conditions: daytime and nighttime. Figure~\ref{fig:sunny} and Figure~\ref{fig:rainy} show the detection result under different weather conditions: sunny weather and rainy weather. We could find that RetinaNet with focal loss perform well in all these different environment situations.

\section{CONCLUSIONS}

In this study, we categorized the latest research of Convolutional Neural Network based object detectors into two groups: one-stage object detector and two-stage object detector. As showed by the experiment result, RetinaNet, a one-stage object detector has proved to be able to achieve state-of-the-art performance for vehicle detection compared with other two-stage object detectors.The incorporated focal loss function,which resolve the critical class imbalance issues of normal one-stage object detectors, give rise to the performance boost. More vehicle based patterns may be explored to improve the one-stage object detector further.

\bibliographystyle{IEEEtran}
\bibliography{citations}

\begin{thebibliography}{10}
\providecommand{\url}[1]{#1}
\csname url@samestyle\endcsname
\providecommand{\newblock}{\relax}
\providecommand{\bibinfo}[2]{#2}
\providecommand{\BIBentrySTDinterwordspacing}{\spaceskip=0pt\relax}
\providecommand{\BIBentryALTinterwordstretchfactor}{4}
\providecommand{\BIBentryALTinterwordspacing}{\spaceskip=\fontdimen2\font plus
\BIBentryALTinterwordstretchfactor\fontdimen3\font minus
  \fontdimen4\font\relax}
\providecommand{\BIBforeignlanguage}[2]{{%
\expandafter\ifx\csname l@#1\endcsname\relax
\typeout{** WARNING: IEEEtran.bst: No hyphenation pattern has been}%
\typeout{** loaded for the language `#1'. Using the pattern for}%
\typeout{** the default language instead.}%
\else
\language=\csname l@#1\endcsname
\fi
#2}}
\providecommand{\BIBdecl}{\relax}
\BIBdecl

\bibitem{dollar2014fast}
P.~Doll{\'a}r, R.~Appel, S.~Belongie, and P.~Perona, ``Fast feature pyramids
  for object detection,'' \emph{IEEE Transactions on Pattern Analysis and
  Machine Intelligence}, vol.~36, no.~8, pp. 1532--1545, 2014.

\bibitem{felzenszwalb2010object}
P.~F. Felzenszwalb, R.~B. Girshick, D.~McAllester, and D.~Ramanan, ``Object
  detection with discriminatively trained part-based models,'' \emph{IEEE
  transactions on pattern analysis and machine intelligence}, vol.~32, no.~9,
  pp. 1627--1645, 2010.

\bibitem{yang2015large}
L.~Yang, P.~Luo, C.~Change~Loy, and X.~Tang, ``A large-scale car dataset for
  fine-grained categorization and verification,'' in \emph{Proceedings of the
  IEEE Conference on Computer Vision and Pattern Recognition}, 2015, pp.
  3973--3981.

\bibitem{rujikietgumjorn2017vehicle}
S.~Rujikietgumjorn and N.~Watcharapinchai, ``Vehicle detection with sub-class
  training using r-cnn for the ua-detrac benchmark,'' in \emph{Advanced Video
  and Signal Based Surveillance (AVSS), 2017 14th IEEE International Conference
  on}.\hskip 1em plus 0.5em minus 0.4em\relax IEEE, 2017, pp. 1--5.

\bibitem{zhou2016image}
Y.~Zhou, H.~Nejati, T.-T. Do, N.-M. Cheung, and L.~Cheah, ``Image-based vehicle
  analysis using deep neural network: A systematic study,'' in \emph{Digital
  Signal Processing (DSP), 2016 IEEE International Conference on}.\hskip 1em
  plus 0.5em minus 0.4em\relax IEEE, 2016, pp. 276--280.

\bibitem{lin2017focal}
T.-Y. Lin, P.~Goyal, R.~Girshick, K.~He, and P.~Doll{\'a}r, ``Focal loss for
  dense object detection,'' \emph{arXiv preprint arXiv:1708.02002}, 2017.

\bibitem{wen2015detrac}
L.~Wen, D.~Du, Z.~Cai, Z.~Lei, M.~Chang, H.~Qi, J.~Lim, M.-H. Yang, and S.~Lyu,
  ``Detrac: A new benchmark and protocol for multi-object tracking,''
  \emph{arXiv preprint arXiv:1511.04136}, 2015.

\bibitem{lecun1989backpropagation}
Y.~LeCun, B.~Boser, J.~S. Denker, D.~Henderson, R.~E. Howard, W.~Hubbard, and
  L.~D. Jackel, ``Backpropagation applied to handwritten zip code
  recognition,'' \emph{Neural computation}, vol.~1, no.~4, pp. 541--551, 1989.

\bibitem{vaillant1994original}
R.~Vaillant, C.~Monrocq, and Y.~Le~Cun, ``Original approach for the
  localisation of objects in images,'' \emph{IEE Proceedings-Vision, Image and
  Signal Processing}, vol. 141, no.~4, pp. 245--250, 1994.

\bibitem{viola2001rapid}
P.~Viola and M.~Jones, ``Rapid object detection using a boosted cascade of
  simple features,'' in \emph{Computer Vision and Pattern Recognition, 2001.
  CVPR 2001. Proceedings of the 2001 IEEE Computer Society Conference on},
  vol.~1.\hskip 1em plus 0.5em minus 0.4em\relax IEEE, 2001, pp. I--I.

\bibitem{dollar2009integral}
P.~Doll{\'a}r, Z.~Tu, P.~Perona, and S.~Belongie, ``Integral channel
  features,'' 2009.

\bibitem{dalal2005histograms}
N.~Dalal and B.~Triggs, ``Histograms of oriented gradients for human
  detection,'' in \emph{Computer Vision and Pattern Recognition, 2005. CVPR
  2005. IEEE Computer Society Conference on}, vol.~1.\hskip 1em plus 0.5em
  minus 0.4em\relax IEEE, 2005, pp. 886--893.

\bibitem{felzenszwalb2010cascade}
P.~F. Felzenszwalb, R.~B. Girshick, and D.~McAllester, ``Cascade object
  detection with deformable part models,'' in \emph{Computer vision and pattern
  recognition (CVPR), 2010 IEEE conference on}.\hskip 1em plus 0.5em minus
  0.4em\relax IEEE, 2010, pp. 2241--2248.

\bibitem{everingham2010pascal}
M.~Everingham, L.~Van~Gool, C.~K. Williams, J.~Winn, and A.~Zisserman, ``The
  pascal visual object classes (voc) challenge,'' \emph{International journal
  of computer vision}, vol.~88, no.~2, pp. 303--338, 2010.

\bibitem{krizhevsky2012imagenet}
A.~Krizhevsky, I.~Sutskever, and G.~E. Hinton, ``Imagenet classification with
  deep convolutional neural networks,'' in \emph{Advances in neural information
  processing systems}, 2012, pp. 1097--1105.

\bibitem{uijlings2013selective}
J.~R. Uijlings, K.~E. Van De~Sande, T.~Gevers, and A.~W. Smeulders, ``Selective
  search for object recognition,'' \emph{International journal of computer
  vision}, vol. 104, no.~2, pp. 154--171, 2013.

\bibitem{girshick2014rich}
R.~Girshick, J.~Donahue, T.~Darrell, and J.~Malik, ``Rich feature hierarchies
  for accurate object detection and semantic segmentation,'' in
  \emph{Proceedings of the IEEE conference on computer vision and pattern
  recognition}, 2014, pp. 580--587.

\bibitem{he2015spatial}
K.~He, X.~Zhang, S.~Ren, and J.~Sun, ``Spatial pyramid pooling in deep
  convolutional networks for visual recognition,'' \emph{IEEE transactions on
  pattern analysis and machine intelligence}, vol.~37, no.~9, pp. 1904--1916,
  2015.

\bibitem{girshick2015fast}
R.~Girshick, ``Fast r-cnn,'' in \emph{Proceedings of the IEEE international
  conference on computer vision}, 2015, pp. 1440--1448.

\bibitem{ren2017faster}
S.~Ren, K.~He, R.~Girshick, and J.~Sun, ``Faster r-cnn: Towards real-time
  object detection with region proposal networks,'' \emph{IEEE transactions on
  pattern analysis and machine intelligence}, vol.~39, no.~6, pp. 1137--1149,
  2017.

\bibitem{shrivastava2016beyond}
A.~Shrivastava, R.~Sukthankar, J.~Malik, and A.~Gupta, ``Beyond skip
  connections: Top-down modulation for object detection,'' \emph{arXiv preprint
  arXiv:1612.06851}, 2016.

\bibitem{lin2016feature}
T.-Y. Lin, P.~Doll{\'a}r, R.~Girshick, K.~He, B.~Hariharan, and S.~Belongie,
  ``Feature pyramid networks for object detection,'' \emph{arXiv preprint
  arXiv:1612.03144}, 2016.

\bibitem{dai2016r}
J.~Dai, Y.~Li, K.~He, and J.~Sun, ``R-fcn: Object detection via region-based
  fully convolutional networks,'' in \emph{Advances in neural information
  processing systems}, 2016, pp. 379--387.

\bibitem{he2017mask}
K.~He, G.~Gkioxari, P.~Doll{\'a}r, and R.~Girshick, ``Mask r-cnn,'' \emph{arXiv
  preprint arXiv:1703.06870}, 2017.

\bibitem{dai2017deformable}
J.~Dai, H.~Qi, Y.~Xiong, Y.~Li, G.~Zhang, H.~Hu, and Y.~Wei, ``Deformable
  convolutional networks,'' \emph{arXiv preprint arXiv:1703.06211}, 2017.

\bibitem{sermanet2013overfeat}
P.~Sermanet, D.~Eigen, X.~Zhang, M.~Mathieu, R.~Fergus, and Y.~LeCun,
  ``Overfeat: Integrated recognition, localization and detection using
  convolutional networks,'' \emph{arXiv preprint arXiv:1312.6229}, 2013.

\bibitem{fu2017dssd}
C.-Y. Fu, W.~Liu, A.~Ranga, A.~Tyagi, and A.~C. Berg, ``Dssd: Deconvolutional
  single shot detector,'' \emph{arXiv preprint arXiv:1701.06659}, 2017.

\bibitem{liu2016ssd}
W.~Liu, D.~Anguelov, D.~Erhan, C.~Szegedy, S.~Reed, C.-Y. Fu, and A.~C. Berg,
  ``Ssd: Single shot multibox detector,'' in \emph{European conference on
  computer vision}.\hskip 1em plus 0.5em minus 0.4em\relax Springer, 2016, pp.
  21--37.

\bibitem{redmon2016you}
J.~Redmon, S.~Divvala, R.~Girshick, and A.~Farhadi, ``You only look once:
  Unified, real-time object detection,'' in \emph{Proceedings of the IEEE
  Conference on Computer Vision and Pattern Recognition}, 2016, pp. 779--788.

\bibitem{redmon2016yolo9000}
J.~Redmon and A.~Farhadi, ``Yolo9000: better, faster, stronger,'' \emph{arXiv
  preprint arXiv:1612.08242}, 2016.

\bibitem{huang2016speed}
J.~Huang, V.~Rathod, C.~Sun, M.~Zhu, A.~Korattikara, A.~Fathi, I.~Fischer,
  Z.~Wojna, Y.~Song, S.~Guadarrama \emph{et~al.}, ``Speed/accuracy trade-offs
  for modern convolutional object detectors,'' \emph{arXiv preprint
  arXiv:1611.10012}, 2016.

\bibitem{sung1996learning}
K.-K. Sung, ``Learning and example selection for object and pattern
  detection,'' 1996.

\bibitem{shrivastava2016training}
A.~Shrivastava, A.~Gupta, and R.~Girshick, ``Training region-based object
  detectors with online hard example mining,'' in \emph{Proceedings of the IEEE
  Conference on Computer Vision and Pattern Recognition}, 2016, pp. 761--769.

\bibitem{rota2017loss}
S.~Rota~Bulo, G.~Neuhold, and P.~Kontschieder, ``Loss max-pooling for semantic
  image segmentation,'' in \emph{Proceedings of the IEEE Conference on Computer
  Vision and Pattern Recognition}, 2017, pp. 2126--2135.

\bibitem{lin2014microsoft}
T.-Y. Lin, M.~Maire, S.~Belongie, J.~Hays, P.~Perona, D.~Ramanan,
  P.~Doll{\'a}r, and C.~L. Zitnick, ``Microsoft coco: Common objects in
  context,'' in \emph{European conference on computer vision}.\hskip 1em plus
  0.5em minus 0.4em\relax Springer, 2014, pp. 740--755.

\bibitem{he2016deep}
K.~He, X.~Zhang, S.~Ren, and J.~Sun, ``Deep residual learning for image
  recognition,'' in \emph{Proceedings of the IEEE conference on computer vision
  and pattern recognition}, 2016, pp. 770--778.

\bibitem{chen2015mxnet}
T.~Chen, M.~Li, Y.~Li, M.~Lin, N.~Wang, M.~Wang, T.~Xiao, B.~Xu, C.~Zhang, and
  Z.~Zhang, ``Mxnet: A flexible and efficient machine learning library for
  heterogeneous distributed systems,'' \emph{arXiv preprint arXiv:1512.01274},
  2015.

\end{thebibliography}

\end{document}